\documentclass[11pt,a4paper]{article}
\usepackage[hyperref]{acl2018}
\usepackage{times}
\usepackage{latexsym}

\usepackage{url}

\def\figref#1{Fig.~\ref{#1}}
\def\secref#1{Sec.~\ref{#1}}

\usepackage{xcolor}
\usepackage{amsmath}
\usepackage{latexsym}
\usepackage{tabularx}
\usepackage{graphicx}
\graphicspath{ {figs/} }

\definecolor{dollarbill}{rgb}{0.52, 0.73, 0.4}

\aclfinalcopy 

\newcommand\blfootnote[1]{%
  \begingroup
  \renewcommand\thefootnote{}\footnote{#1}%
  \addtocounter{footnote}{-1}%
  \endgroup
}

\title{Soft Layer-Specific Multi-Task Summarization \\ with Entailment and Question Generation}

\author{Han Guo$^*$ \;\;\;\;\;\;\; Ramakanth Pasunuru$^*$ \;\;\;\;\;\;\; Mohit Bansal \\
  UNC Chapel Hill \\
  {\tt \{hanguo, ram, mbansal\}@cs.unc.edu} \\
 }

\date{}

\begin{document}
\maketitle

\begin{abstract}
An accurate abstractive summary of a document should contain all its salient information and should be logically entailed by the input document. We improve these important aspects of abstractive summarization via multi-task learning with the auxiliary tasks of question generation and entailment generation, where the former teaches the summarization model how to look for salient questioning-worthy details, and the latter teaches the model how to rewrite a summary which is a directed-logical subset of the input document. We also propose novel multi-task architectures with high-level (semantic) layer-specific sharing across multiple encoder and decoder layers of the three tasks, as well as soft-sharing mechanisms (and show performance ablations and analysis examples of each contribution). Overall, we achieve statistically significant improvements over the state-of-the-art on both the CNN/DailyMail and Gigaword datasets, as well as on the DUC-2002 transfer setup. We also present several quantitative and qualitative analysis studies of our model's learned saliency and entailment skills. 

\end{abstract}

\section{Introduction}
\label{sec-intro}

Abstractive summarization is the challenging NLG task of compressing and rewriting a document into a short, relevant, salient, and coherent summary.\blfootnote{$*$ Equal contribution (published at ACL 2018).}
It has numerous applications such as summarizing storylines, event understanding, etc. As compared to extractive or compressive summarization~\cite{jing2000sentence,knight2002summarization,clarke2008global,filippova2015sentence,henss2015reinforcement}, abstractive summaries are based on rewriting as opposed to selecting. Recent end-to-end, neural sequence-to-sequence models and larger datasets have allowed substantial progress on the abstractive task, with ideas ranging from copy-pointer mechanism and redundancy coverage, to metric reward based reinforcement learning~\cite{rush2015neural,chopra2016abstractive,nallapati2016abstractive,see2017get}.

Despite these strong recent advancements, there is still a lot of scope for improving the summary quality generated by these models. A good rewritten summary is one that contains all the salient information from the document, is logically followed (entailed) by it, and avoids redundant information. The redundancy aspect was addressed by coverage models~\cite{Suzuki2016Summ,Chen2016DistractionBasedNN,nallapati2016abstractive,see2017get}, but we still need to teach these models about how to better detect salient information from the input document, as well as about better logically-directed natural language inference skills. 

In this work, we improve abstractive text summarization via soft, high-level (semantic) layer-specific multi-task learning with two relevant auxiliary tasks. The first is that of document-to-question generation, which teaches the summarization model about what are the right questions to ask, which in turn is directly related to what the salient information in the input document is. The second auxiliary task is a premise-to-entailment generation task to teach it how to rewrite a summary which is a directed-logical subset of (i.e., logically follows from) the input document, and contains no contradictory or unrelated information. For the question generation task, we use the SQuAD dataset~\cite{rajpurkar2016squad}, where we learn to generate a question given a sentence containing the answer, similar to the recent work by~\newcite{du2017learning}. Our entailment generation task is based on the recent SNLI classification dataset and task~\cite{bowman2015large}, converted to a generation task~\cite{pasunuru2017multitask}.  

Further, we also present novel multi-task learning architectures based on multi-layered encoder and decoder models, where we empirically show that it is substantially better to share the higher-level semantic layers between the three aforementioned tasks, while keeping the lower-level (lexico-syntactic) layers unshared. We also explore different ways to optimize the shared parameters and show that `soft' parameter sharing achieves higher performance than hard sharing. 

Empirically, our soft, layer-specific sharing model with the question and entailment generation auxiliary tasks achieves statistically significant improvements over the state-of-the-art on both the CNN/DailyMail and Gigaword datasets. It also performs significantly better on the DUC-2002 transfer setup, demonstrating its strong generalizability as well as the importance of auxiliary knowledge in low-resource scenarios. We also report improvements on our auxiliary question and entailment generation tasks over their respective previous state-of-the-art.
Moreover, we significantly decrease the training time of the multi-task models by initializing the individual tasks from their pretrained baseline models. 
Finally, we present human evaluation studies as well as detailed quantitative and qualitative analysis studies of the improved saliency detection and logical inference skills learned by our multi-task model.


\section{Related Work}
\label{Related Work}

Automatic text summarization has been progressively improving over the time, initially more focused on extractive and compressive models~\cite{jing2000sentence,knight2002summarization,clarke2008global,filippova2015sentence,kedzie2015predicting}, and moving more towards compressive and abstractive summarization based on graphs and concept maps~\cite{giannakopoulos2009automatic,ganesan2010opinosis,falke2017bringing} and discourse trees~\cite{gerani2014abstractive}, syntactic parse trees~\cite{cheung2014unsupervised,wang2016sentence}, and Abstract Meaning Representations (AMR)~\cite{liu2015toward,dohare2017text}.  
Recent work has also adopted machine translation inspired neural seq2seq models for abstractive summarization with advances in hierarchical, distractive, saliency, and graph-attention modeling ~\cite{rush2015neural,chopra2016abstractive,nallapati2016abstractive,Chen2016DistractionBasedNN,Tan2017AbstractiveDS}.~\newcite{paulus2017deep} and~\newcite{henss2015reinforcement} incorporated recent advances from reinforcement learning. Also,~\newcite{see2017get} further improved results via pointer-copy mechanism and addressed the redundancy with coverage mechanism.

Multi-task learning (MTL) is a useful paradigm to improve the generalization performance of a task with related tasks while sharing some common parameters/representations~\cite{caruana1998multitask,argyriou2007multi,kumar2012learning}. 
Several recent works have adopted MTL in neural models~\cite{luong2015multi,Misra2016CrossStitchNF,Hashimoto2017AJM,pasunuru2017multitask,Ruder2017SluiceNL,Kaiser2017OneMT}. 
Moreover, some of the above works have investigated the use of shared vs unshared sets of parameters. On the other hand, we investigate the importance of soft parameter sharing and high-level versus low-level layer-specific sharing.

Our previous workshop paper~\cite{Pasunuru2017TowardsIA} presented some preliminary results for multi-task learning of textual summarization with entailment generation. This current paper has several major differences: 
(1) We present question generation as an additional effective auxiliary task to enhance the important complementary aspect of saliency detection;
(2) Our new high-level layer-specific sharing approach is significantly better than alternative layer-sharing approaches (including the decoder-only sharing by ~\newcite{Pasunuru2017TowardsIA});
(3) Our new soft sharing parameter approach gives stat. significant improvements over hard sharing;
(4) We propose a useful idea of starting multi-task models from their pretrained baselines, which significantly speeds up our experiment cycle\footnote{About $4$-$5$ days for~\newcite{Pasunuru2017TowardsIA} approach vs. only $10$ hours for us. This will allow the community to try many more multi-task training and tuning ideas faster.};
(5) For evaluation, we show diverse improvements of our soft, layer-specific MTL model (over state-of-the-art pointer+coverage baselines) on the CNN/DailyMail, Gigaword, as well as DUC datasets; we also report human evaluation plus analysis examples of learned saliency and entailment skills; we also report improvements on the auxiliary question and entailment generation tasks over their respective previous state-of-the-art.

In our work, we use a question generation task to improve the saliency of abstractive summarization in a multi-task setting. Using the SQuAD dataset~\cite{rajpurkar2016squad}, we learn to generate a question given the sentence containing the answer span in the comprehension (similar to~\newcite{du2017learning}).
For the second auxiliary task of entailment generation, we use the generation version of the RTE classification task~\cite{dagan2006pascal,lai2014illinois,jimenez2014unal,bowman2015large}. Some previous work has explored the use of RTE for redundancy detection in summarization by modeling graph-based relationships between sentences to select the most non-redundant sentences~\cite{mehdad2013abstractive,gupta2014text}, whereas our approach is based on multi-task learning.
%

\section{Models}

First, we introduce our pointer+coverage baseline model and then our two auxiliary tasks: question generation and entailment generation (and finally the multi-task learning models in Sec.~\ref{sec:multitask-learning}).

\subsection{Baseline Pointer+Coverage Model}
\label{subsec:attention-model}
We use a sequence-attention-sequence model with a 2-layer bidirectional LSTM-RNN encoder and a 2-layer uni-directional LSTM-RNN decoder, along with~\newcite{bahdanau2014neural} style attention. Let $x=\{x_1,x_2,...,x_m\}$ be the source document and $y=\{y_1,y_2,...,y_n\}$ be the target summary. The output summary generation vocabulary distribution conditioned over the input source document is $P_v(y|x;\theta) = \prod_{t=1}^{n}p_v(y_t|y_{1:t-1},x;\theta)$.
Let the decoder hidden state be $s_t$ at time step $t$ and let $c_t$ be the context vector which is defined as a weighted combination of encoder hidden states. We concatenate the decoder's (last) RNN layer hidden state $s_t$ and context vector $c_t$ and apply a linear transformation, and then project to the vocabulary space by another linear transformation. Finally, the conditional vocabulary distribution at each time step $t$ of the decoder is defined as: 
\begin{equation}
\begin{small}
\label{eq:vocab-dist}
p_v(y_t|y_{1:t-1},x;\theta) = \mathrm{sfm} (V_p(W_f[s_t;c_t]+b_f)+b_p)
\end{small}
\end{equation}
where, $W_f$, $V_p$, $b_f$, $b_p$ are trainable parameters, and $\mathrm{sfm(\cdot)}$ is the softmax function. 

\paragraph{Pointer-Generator Networks}
\label{subsec:pointer-model}
Pointer mechanism~\cite{vinyals2015pointer} helps in directly copying the words from the source sequence during target sequence generation, which is a good fit for a task like summarization. Our pointer mechanism approach is similar to~\newcite{see2017get}, who use a soft switch based on the generation probability $p_g = \sigma(W_g c_t+U_g s_t + V_g e_{w_{t-1}} + b_g)$, where $\sigma (\cdot)$ is a sigmoid function, $W_g$, $U_g$, $V_g$ and $b_g$ are parameters learned during training. $e_{w_{t-1}}$ is the previous time step output word embedding. The final word distribution is $P_f(y) = p_g \cdot P_v(y) + (1-p_g) \cdot P_c(y)$, where $P_v$ vocabulary distribution is as shown in Eq.~\ref{eq:vocab-dist}, and copy distribution $P_c$ is based on the attention distribution over source document words. 
%
\paragraph{Coverage Mechanism}
\label{subsec:coverage-mechanisml}
Following previous work~\cite{see2017get}, coverage helps alleviate the issue of word repetition while generating long summaries. We maintain a coverage vector $\hat{c}_t = \sum_{t=0}^{t-1}\alpha_t$ that sums over all of the previous time steps attention distributions $\alpha_t$, and this is added as input to the attention mechanism. Coverage loss is $L_{cov}(\theta) = \sum_t\sum_i min(\alpha_{t,i},\hat{c}_{t,i})$. 
Finally, the total loss is a weighted combination of cross-entropy loss and coverage loss: 
\vspace{-5pt}
\begin{equation}
L(\theta) = -\log P_f(y) + \lambda L_{cov}(\theta)
\vspace{-5pt}
\end{equation}
where $\lambda$ is a tunable hyperparameter.

\subsection{Two Auxiliary Tasks}
Despite the strengths of the baseline model described above with attention, pointer, and coverage, a good summary should also contain maximal salient information and be a directed logical entailment of the source document. We teach these skills to the abstractive summarization model via multi-task training with two related auxiliary tasks: question generation task and entailment generation.
\paragraph{Question Generation}
\label{paragraph:question-generation}
The task of question generation is to generate a question from a given input sentence, which in turn is related to the skill of being able to find the important salient information to ask questions about. First the model has to identify the important information present in the given sentence, then it has to frame (generate) a question based on this salient information, such that, given the sentence and the question, one has to be able to predict the correct answer (salient information in this case). A good summary should also be able to find and extract all the salient information in the given source document, and hence we incorporate such capabilities into our abstractive text summarization model by multi-task learning it with a question generation task, sharing some common parameters/representations (see more details in Sec.~\ref{sec:multitask-learning}).
For setting up the question generation task, we follow~\newcite{du2017learning} and use the SQuAD dataset to extract sentence-question pairs. Next, we use the same sequence-to-sequence model architecture as our summarization model. 
Note that even though our question generation task is generating one question at a time\footnote{We also tried to generate all the questions at once from the full document, but we obtained low accuracy because of this task's challenging nature and overall less training data.}, our multi-task framework (see Sec.~\ref{sec:multitask-learning}) is set up in such a way that the sentence-level knowledge from this auxiliary task can help the document-level primary (summarization) task to generate multiple salient facts -- by sharing high-level semantic layer representations. See Sec.~\ref{para:saliency-detection} and Table~\ref{table:saliency_results_2} for a quantitative evaluation showing that the multi-task model can find multiple (and more) salient phrases in the source document. Also see Sec.~\ref{sec:analysis} (and supp) for challenging qualitative examples where baseline and SotA models only recover a small subset of salient information but our multi-task model with question generation is able to detect more of the important information.

\paragraph{Entailment Generation}
The task of entailment generation is to generate a hypothesis which is entailed by (or logically follows from) the given premise as input. 
In summarization, the generation decoder also needs to generate a summary that is entailed by the source document, i.e., does not contain any contradictory or unrelated/extraneous information as compared to the input document. We again incorporate such inference capabilities into the summarization model via multi-task learning, sharing some common representations/parameters between our summarization and entailment generation model (more details in Sec.~\ref{sec:multitask-learning}).
For this task, we use the entailment-labeled pairs from the SNLI dataset~\cite{bowman2015large} and set it up as a generation task (using the same strong model architecture as our abstractive summarization model). 
See Sec.~\ref{para:qualitative-analysis} and Table~\ref{table:entailment_analysis_results} for a quantitative evaluation showing that the multi-task model is better entailed by the source document and has fewer extraneous facts.
Also see Sec.~\ref{para:qualitative-analysis} and supplementary for qualitative examples of how our multi-task model with the entailment auxiliary task is able to generate more logically-entailed summaries than the baseline and SotA models, which instead produce extraneous, unrelated words not present (in any paraphrased form) in the source document.
%

\begin{figure}
\centering
\includegraphics[width=0.98\linewidth]{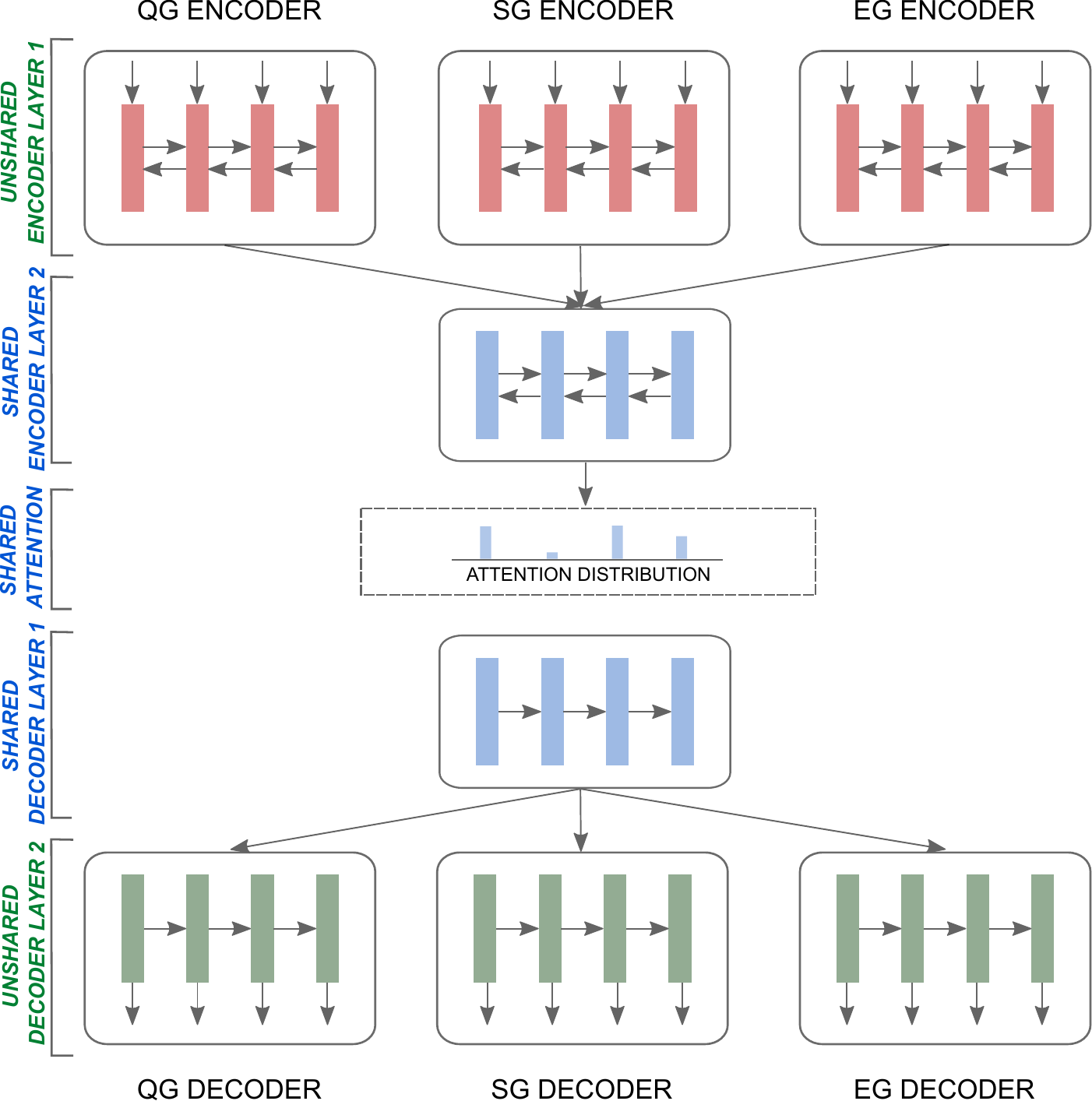}
\vspace{-10pt}
\caption{Overview of our multi-task model with parallel training of three tasks: abstractive summary generation (SG), question generation (QG), and entailment generation (EG). We share the `blue' color representations across all the three tasks, i.e., second layer of encoder, attention parameters, and first layer of decoder.}
\vspace{-10pt}
\label{fig:mutlitask model}
\end{figure}
\section{Multi-Task Learning}
\label{sec:multitask-learning}
We employ multi-task learning for parallel training of our three tasks: abstractive summarization, question generation, and entailment generation. In this section, we describe our novel layer-specific, soft-sharing approaches and other multi-task learning details.

\subsection{Layer-Specific Sharing Mechanism}
\label{subsec:sharing-mechanism}

Simply sharing all parameters across the related tasks is not optimal, because models for different tasks have different input and output distributions, esp. for low-level vs. high-level parameters. Therefore, related tasks should share some common representations (e.g., high-level information), as well as need their own individual task-specific representations (esp. low-level information). To this end, we allow different components of model parameters of related tasks to be shared vs. unshared, as described next.

\noindent\textbf{Encoder Layer Sharing}:
~\newcite{belinkov2017} observed that lower layers (i.e., the layers closer to the input words) of RNN cells in a seq2seq machine translation model learn to represent word structure, while higher layers (farther from input) are more focused on high-level semantic meanings (similar to findings in the computer vision community for image features~\cite{zeiler2014visualizing}). We believe that while textual summarization, question generation, and entailment generation have different training data distributions and low-level representations, they can still benefit from sharing their models' high-level components (e.g., those that capture the skills of saliency and inference). Thus, we keep the lower-level layer (i.e., first layer closer to input words) of the 2-layer encoder of all three tasks unshared, while we share the higher layer (second layer in our model as shown in Fig.~\ref{fig:mutlitask model}) across the three tasks. 

\noindent\textbf{Decoder Layer Sharing}:
Similarly for the decoder, lower layers (i.e., the layers closer to the output words) learn to represent word structure for generation, while higher layers (farther from output) are more focused on high-level semantic meaning. Hence, we again share the higher level components (first layer in the decoder far from output as shown in Fig.~\ref{fig:mutlitask model}), while keeping the lower layer (i.e., second layer) of decoders of all three tasks unshared.

\noindent\textbf{Attention Sharing}: As described in Sec.~\ref{subsec:attention-model}, the attention mechanism defines an attention distribution over high-level layer encoder hidden states and since we share the second, high-level (semantic) layer of all the encoders, it is intuitive to share the attention parameters as well. 

\subsection{Soft vs. Hard Parameter Sharing}
\label{subsec:soft-vs-hard}

\noindent\textbf{Hard-sharing}: In the most common multi-task learning hard-sharing approach, the parameters to be shared are forced to be the same. As a result, gradient information from multiple tasks will directly pass through shared parameters, hence forcing a common space representation for all the related tasks.
\noindent\textbf{Soft-sharing}:
In our soft-sharing approach, we encourage shared parameters to be close in representation space by penalizing their $l_2$ distances. Unlike hard sharing, this approach gives more flexibility for the tasks by only loosely coupling the shared space representations. We minimize the following loss function for the primary task in soft-sharing approach:
\vspace{-5pt}
\begin{equation} \label{eq:p_f_mtl}
L(\theta) = - \log P_f(y) + \lambda L_{cov}(\theta) + \gamma \| \theta_{s} - \psi_{s} \|
\vspace{-5pt}
\end{equation}
where $\gamma$ is a hyperparameter, $\theta$ represents the primary summarization task's full parameters, while $\theta_{s}$ and $\psi_{s}$ represent the shared parameter subset between the primary and auxiliary tasks.

\begin{table*}
\begin{center}
\begin{small}
\begin{tabular}{|l|c|c|c|c|c|c|}
\hline
Models & \ \ ROUGE-1 \ \  & \ \ ROUGE-2 \ \ & \ \ ROUGE-L \ \ & \ \ METEOR \ \ \\
\hline
\multicolumn{5}{|c|}{\textsc{Previous Work}}\\
\hline
Seq2Seq(50k vocab)~\cite{see2017get} & 31.33 & 11.81 & 28.83 & 12.03  \\
Pointer~\cite{see2017get} & 36.44 & 15.66 & 33.42 & 15.35 \\
Pointer+Coverage~\cite{see2017get} $\star$ & 39.53 & 17.28 & 36.38 & 18.72  \\
Pointer+Coverage~\cite{see2017get} $\dagger$ & 38.82 & 16.81 & 35.71 & 18.14 \\
\hline
\multicolumn{5}{|c|}{\textsc{Our Models}}\\
\hline
Two-Layer Baseline (Pointer+Coverage) $\otimes$ & 39.56 & 17.52 & 36.36  & 18.17 \\
$\otimes$ + Entailment Generation & 39.84 & 17.63 & 36.54 & 18.61 \\
$\otimes$ + Question Generation & 39.73 & 17.59 & 36.48 & 18.33  \\
$\otimes$ + Entailment Gen. + Question Gen. & 39.81 & 17.64 & 36.54 & 18.54 \\
\hline
\end{tabular}
\end{small}
\end{center}
\vspace{-10pt}
\caption{CNN/DailyMail summarization results. ROUGE scores are full length F-1 (as previous work). 
All the multi-task improvements are statistically significant over the state-of-the-art baseline.
\vspace{-10pt}
}
\label{table:cnndm_results}
\end{table*}

\subsection{Fast Multi-Task Training}
During multi-task learning, we alternate the mini-batch optimization of the three tasks, based on a tunable `mixing ratio' $\alpha_s:\alpha_q:\alpha_e$; i.e., optimizing the summarization task for $\alpha_s$ mini-batches followed by optimizing the question generation task for $\alpha_q$ mini-batches, followed by entailment generation task for $\alpha_e$ mini-batches (and for 2-way versions of this, we only add one auxiliary task at a time). We continue this process  until all the models converge. Also, importantly, instead of training from scratch, we start the primary task (summarization) from a $90\%$-converged model of its baseline to make the training process faster. We observe that starting from a fully-converged baseline makes the model stuck in a local minimum. In addition, we also start all auxiliary models from their $90\%$-converged baselines, as we found that starting the auxiliary models from scratch has a chance to pull the primary model's shared parameters towards randomly-initialized auxiliary model's shared parameters.

\section{Experimental Setup}
\label{sec-setup}

\label{sec:datasets}
\noindent\textbf{Datasets}: We use CNN/DailyMail dataset~\cite{hermann2015teaching,nallapati2016abstractive} and Gigaword~\cite{rush2015neural} datasets for summarization, and the Stanford Natural Language Inference (SNLI) corpus~\cite{bowman2015large} and the Stanford Question Answering Dataset (SQuAD)~\cite{rajpurkar2016squad} datasets for our entailment and question generation tasks, resp. We also show generalizability/transfer results on DUC-2002 with our CNN/DM trained models. Supplementary contains dataset details.

\noindent\textbf{Evaluation Metrics}: We use the standard ROUGE evaluation package~\cite{lin2004rouge} for reporting the results on all of our summarization models. Following previous work~\cite{chopra2016abstractive,nallapati2016abstractive}, we use ROUGE full-length F1 variant for all our results.
Following~\newcite{see2017get}, we also report METEOR~\cite{banerjee2005meteor} using the MS-COCO evaluation script~\cite{chen2015microsoft}.

\noindent\textbf{Human Evaluation Criteria}: We used Amazon MTurk to perform human evaluation of summary \emph{relevance} and \emph{readability}. We selected human annotators that were located in the US, had an approval rate greater than 95\%, and had at least 10,000 approved HITs. For the pairwise model comparisons discussed in~\secref{subsec:human-evaluation}, we showed the annotators the input article, the ground truth summary, and the two model summaries (randomly shuffled to anonymize model identities) -- we then asked them to choose the better among the two model summaries or choose `Not-Distinguishable' if both summaries are equally good/bad.
Instructions for relevance were defined based on the summary containing salient/important information from the given article, being correct (i.e., avoiding contradictory/unrelated information), and avoiding redundancy.
Instructions for readability were based on the summary's fluency, grammaticality, and coherence.

\paragraph{Training Details}
\label{subsec:training-details}
All our soft/hard and layer-specific sharing decisions were made on the validation/development set. Details of RNN hidden state sizes, Adam optimizer, mixing ratios, etc. are provided in the supplementary for reproducibility.

\section{Results}
\label{sec-results}

\subsection{Summarization (Primary Task) Results}

\paragraph{Pointer+Coverage Baseline}
We start from the strong model of~\newcite{see2017get}.\footnote{We use two layers so as to allow our high- versus low-level layer sharing intuition. Note that this does not increase the parameter size much (23M versus 22M for ~\newcite{see2017get}).} Table~\ref{table:cnndm_results} shows that our baseline model performs better than or comparable to~\newcite{see2017get}.\footnote{As mentioned in the github for~\newcite{see2017get}, their publicly released pretrained model produces the lower scores that we represent by $\dagger$ in Table~\ref{table:cnndm_results}.} 
On Gigaword dataset, our baseline model (with pointer only, since coverage not needed for this single-sentence summarization task) performs better than all previous works, as shown in Table~\ref{table:gigaword_results}.

\begin{table}[t]
\begin{center}
\begin{small}
\begin{tabular}{|l|c|c|c|}
\hline
Models & R-1 & R-2 & R-L \\ 
\hline
\multicolumn{4}{|c|}{\textsc{Previous Work}}\\
\hline
ABS+~\cite{rush2015neural} & 29.76 & 11.88 & 26.96 \\ 
RAS-El~\cite{chopra2016abstractive} & 33.78 & 15.97 & 31.15 \\ 
lvt2k~\cite{nallapati2016abstractive} & 32.67 & 15.59 & 30.64 \\ 
\newcite{Pasunuru2017TowardsIA} & 32.75 & 15.35 & 30.82 \\ 
\hline
\multicolumn{4}{|c|}{\textsc{Our Models}}\\
\hline
2-Layer Pointer Baseline $\otimes$ & 34.26 & 16.40 & 32.03 \\ 
$\otimes$ + Entailment Generation& 35.45 & 17.16 & 33.19 \\ 
$\otimes$ + Question Generation& 35.48 & 17.31 & 32.97 \\ 
$\otimes$ + Entailment + Question& 35.98 & 17.76 & 33.63 \\ 
\hline
\end{tabular}
\end{small}
\end{center}
\vspace{-10pt}
\caption{Summarization results on Gigaword. ROUGE scores are full length F-1. All the multi-task improvements are statistically significant over the state-of-the-art baseline.}
\label{table:gigaword_results}
\vspace{-12pt}
\end{table}

\paragraph{Multi-Task with Entailment Generation}
We first perform multi-task learning between abstractive summarization and entailment generation with soft-sharing of parameters as discussed in Sec.~\ref{sec:multitask-learning}.
Table~\ref{table:cnndm_results} and Table~\ref{table:gigaword_results} shows that this multi-task setting is better than our strong baseline models and the improvements are statistically significant on all metrics\footnote{Stat. significance is computed via bootstrap test~\cite{noreen1989computer,efron1994introduction} with 100K samples.} on both CNN/DailyMail ($p<0.01$ in ROUGE-1/ROUGE-L/METEOR and $p<0.05$ in ROUGE-2) and Gigaword ($p<0.01$ on all metrics) datasets, showing that entailment generation task is inducing useful inference skills to the summarization task (also see analysis examples in Sec.~\ref{sec:analysis}).

\paragraph{Multi-Task with Question Generation}
For multi-task learning with question generation, the improvements are statistically significant in ROUGE-1 ($p<0.01$), ROUGE-L ($p<0.05$), and METEOR ($p<0.01$) for CNN/DailyMail and in all metrics ($p<0.01$) for Gigaword, compared to the respective baseline models. Also, Sec.~\ref{sec:analysis} presents quantitative and qualitative analysis of this model's improved saliency.\footnote{In order to verify that our improvements were from the auxiliary tasks' specific character/capabilities and not just due to adding more data, we separately trained word embeddings on each auxiliary dataset (i.e., SNLI and SQuAD) and incorporated them into the summarization model. We found that both our 2-way multi-task models perform significantly better than these models using the auxiliary word-embeddings, suggesting that merely adding more data is not enough.}

\begin{table}[t]
\begin{small}
\begin{center}
\begin{tabular}{|l|c|c|c|}
\hline
Models & Relevance & Readability & Total \\
\hline
\multicolumn{4}{|c|}{\textsc{MTL vs. Baseline}}\\
\hline
MTL wins & 43 & 40 & 83\\
Baseline wins & 22 & 24 & 46 \\
Non-distinguish. & 35 & 36 & 71 \\
\hline
\multicolumn{4}{|c|}{\textsc{MTL vs.~\newcite{see2017get}}}\\
\hline
MTL wins & 39 & 33 & 72 \\
See \shortcite{see2017get} wins & 29 & 38 & 67\\
Non-distinguish. & 32 & 29 & 61\\
\hline
\end{tabular}
\end{center}
\vspace{-10pt}
\caption{CNN/DM Human Evaluation: pairwise comparison between our 3-way multi-task (MTL) model w.r.t. our baseline and~\newcite{see2017get}.
}
\label{table:human-eval-results_new}
\end{small}
\end{table}

\begin{table}[t]
\begin{small}
\begin{center}
\begin{tabular}{|l|c|c|c|}
\hline
Models & Relevance & Readability & Total \\
\hline
MTL wins & 33 & 32 & 65\\
Baseline wins & 22 & 22 & 44 \\
Non-distinguish. & 45 & 46 & 91 \\
\hline
\end{tabular}
\end{center}
\vspace{-10pt}
\caption{Gigaword Human Evaluation: pairwise comparison between our 3-way multi-task (MTL) model w.r.t. our baseline.
}
\label{table:human-eval-results_gigaword}
\vspace{-7pt}
\end{small}
\end{table}

\paragraph{Multi-Task with Entailment and Question Generation}
Finally, we perform multi-task learning with all three tasks together, achieving the best of both worlds (inference skills and saliency). Table~\ref{table:cnndm_results} and Table~\ref{table:gigaword_results} show that our full multi-task model achieves the best scores on CNN/DailyMail and Gigaword datasets, and the improvements are statistically significant on all metrics on both CNN/DailyMail ($p<0.01$ in ROUGE-1/ROUGE-L/METEOR and $p<0.02$ in ROUGE-2) and Gigaword ($p<0.01$ on all metrics). 
Finally, our 3-way multi-task model (with both entailment and question generation) outperforms the publicly-available pretrained result ($\dagger$) of the previous SotA~\cite{see2017get} with stat. significance ($p<0.01$), as well the higher-reported results ($\star$) on ROUGE-1/ROUGE-2 ($p<0.01$).

\subsection{Human Evaluation}
\label{subsec:human-evaluation}

We also conducted a blind human evaluation on Amazon MTurk for relevance and readability, based on 100 samples, for both CNN/DailyMail and Gigaword (see instructions in~\secref{sec-setup}). Table.~\ref{table:human-eval-results_new} shows the CNN/DM results where we do pairwise comparison between our 3-way multi-task model's output summaries w.r.t. our baseline summaries and w.r.t.~\newcite{see2017get} summaries. As shown, our 3-way multi-task model achieves both higher relevance and higher readability scores w.r.t. the baseline. W.r.t.~\newcite{see2017get}, our MTL model is higher in relevance scores but a bit lower in readability scores (and is higher in terms of total aggregate scores). One potential reason for this lower readability score is that our entailment generation auxiliary task encourages our summarization model to rewrite more and to be more abstractive than~\newcite{see2017get} -- see abstractiveness results in Table~\ref{table:abstractiveness_results}. 

We also show human evaluation results on the Gigaword dataset in Table~\ref{table:human-eval-results_gigaword} (again based on pairwise comparisons for 100 samples), where we see that our MTL model is better than our state-of-the-art baseline on both relevance and readability.\footnote{Note that we did not have output files of any previous work's model on Gigaword; however, our baseline is already a strong state-of-the-art model as shown in Table~\ref{table:gigaword_results}.}

\begin{table}[t]
\begin{small}
\begin{center}
\begin{tabular}{|l|c|c|c|}
\hline
Models & R-1 & R-2 & R-L \\
\hline
\citet{see2017get} & 34.30 & 14.25 & 30.82 \\
Baseline & 35.96 & 15.91 & 32.92 \\
Multi-Task (EG + QG) & 36.73 & 16.15 & 33.58  \\
\hline
\end{tabular}
\end{center}
\vspace{-10pt}
\caption{
ROUGE F1 scores on DUC-2002. \vspace{-10pt}}
\label{table:duc2002}
\vspace{-7pt}
\end{small}
\end{table}

\subsection{Generalizability Results (DUC-2002) }
Next, we also tested our model's generalizability/transfer
skills, where we take the models
trained on CNN/DailyMail and directly test them on DUC-2002. We take our baseline and 3-way multi-task models, plus the pointer-coverage model from~\citet{see2017get}.\footnote{We use the publicly-available pretrained model from~\citet{see2017get}'s github for these DUC transfer results, which produces the $\dagger$ results in Table~\ref{table:cnndm_results}. All other comparisons and analysis in our paper are based on their higher $\star$ results.} We only re-tune the beam-size for each of these three models separately (based on DUC-2003 as the validation set).\footnote{We follow previous work which has shown that larger beam values are better and feasible for DUC corpora. However, our MTL model still achieves stat. significant improvements ($p<0.01$ in all metrics) over~\citet{see2017get} without beam retuning (i.e., with beam $= 4$).} As shown in Table~\ref{table:duc2002}, our multi-task model achieves statistically significant improvements over the strong baseline ($p<0.01$ in ROUGE-1 and ROUGE-L) and the pointer-coverage model from~\newcite{see2017get} ($p<0.01$ in all metrics). This demonstrates that our model is able to generalize well and that the auxiliary knowledge helps more in low-resource scenarios.

\subsection{Auxiliary Task Results}
\label{subsec:auxiliary-performance}
In this section, we discuss the individual/separated performance of our auxiliary tasks. 

\begin{table}[t]
\begin{small}
\begin{center}
\begin{tabular}{|l|c|c|c|c|}
\hline
Models & M & C & R & B \\
\hline
Pasunuru\&Bansal~\shortcite{pasunuru2017multitask} & 29.6 & 117.8 & 62.4 & 40.6 \\
Our 1-layer pointer EG & 32.4 & 139.3 & 65.1 & 43.6 \\
Our 2-layer pointer EG & 32.3 & 140.0 & 64.4 & 43.7 \\
\hline
\end{tabular}
\end{center}
\vspace{-10pt}
\caption{Performance of our pointer-based entailment generation (EG) models compared with previous SotA work. M, C, R, B are short for Meteor, CIDEr-D, ROUGE-L, and BLEU-4, resp.}
\label{table:entailment_results}
\end{small}
\end{table}

\begin{table}[t]
\begin{small}
\begin{center}
\begin{tabular}{|l|c|c|c|c|}
\hline
Models & M & C & R & B \\
\hline
\newcite{du2017learning} & 15.2 & - & 38.0 & 10.8 \\
Our 1-layer pointer QG & 15.4 & 75.3 & 36.2 & 9.2 \\
Our 2-layer pointer QG & 17.5 & 95.3 & 40.1 & 13.8 \\
\hline
\end{tabular}
\end{center}
\vspace{-10pt}
\caption{Performance of our pointer-based question generation (QG) model w.r.t. previous work.}
\label{table:question_generation_results}
\vspace{-7pt}
\end{small}
\end{table}

\paragraph{Entailment Generation}
We use the same architecture as described in Sec.~\ref{subsec:attention-model} with pointer mechanism, and Table~\ref{table:entailment_results} compares our model's performance to~\newcite{pasunuru2017multitask}. Our pointer mechanism gives a performance boost, since the entailment generation task involves copying from the given premise sentence, whereas the 2-layer model seems comparable to the 1-layer model. Also, the supplementary shows some output examples from our entailment generation model. 

\paragraph{Question Generation}
Again, we use same architecture as described in Sec.~\ref{subsec:attention-model} along with pointer mechanism for the task of question generation. Table~\ref{table:question_generation_results} compares the performance of our model w.r.t. the state-of-the-art~\newcite{du2017learning}. Also, the supplementary shows some output examples from our question generation model.


\begin{table}[t]
\begin{center}
\begin{small}
\begin{tabular}{|l|c|c|c|c|c|}
\hline
Models & R-1 & R-2 & R-L & M\\
\hline
\textbf{Final Model} & \textbf{39.81} & \textbf{17.64} & \textbf{36.54} & \textbf{18.54} \\
\hline
\multicolumn{5}{|c|}{\textsc{Soft-vs.-Hard Sharing}}\\
\hline
Hard-sharing & 39.51 & 17.44 & 36.33 & 18.21 \\
\hline
\multicolumn{5}{|c|}{\textsc{Layer Sharing Methods}}\\
\hline
D1+D2 & 39.62 & 17.49 & 36.44 & 18.34 \\
E1+D2 & 39.51 & 17.51 & 36.37 & 18.15 \\
E1+Attn+D2 & 39.32 & 17.36 & 36.11 & 17.88 \\
\hline
\end{tabular}
\end{small}
\end{center}
\vspace{-10pt}
\caption{Ablation studies comparing our final multi-task model with hard-sharing and different alternative layer-sharing methods. Here E1, E2, D1, D2, Attn refer to parameters of the first/second layer of encoder/decoder, and attention parameters. Improvements of final model upon ablation experiments are all stat. signif. with
$p<0.05$.
}
\label{table:ablation-studies}
\end{table}

\begin{table}[t]
\begin{small}
\begin{center}
\begin{tabular}{|l|c|}
\hline
Models & Average Entailment Probability \\
\hline
Baseline &  0.907 \\
Multi-Task (EG) & 0.912 \\
\hline
\end{tabular}
\end{center}
\vspace{-10pt}
\caption{Entailment classification results of our baseline vs. EG-multi-task model ($p<0.001$).}
\label{table:entailment_analysis_results}
\vspace{-5pt}
\end{small}
\end{table}

\begin{table}[t]
\begin{small}
\begin{center}
\begin{tabular}{|l|c|}
\hline
Models & Average Match Rate \\
\hline
Baseline & 27.75 \% \\
Multi-Task (QG) & 28.06 \% \\
\hline
\end{tabular}
\end{center}
\vspace{-10pt}
\caption{Saliency classification results of our baseline vs. QG-multi-task model ($p<0.01$).}
\label{table:saliency_results_2}
\vspace{-4pt}
\end{small}
\end{table}

\section{Ablation and Analysis Studies}
\label{sec:analysis}

\paragraph{Soft-sharing vs. Hard-sharing}
As described in Sec.~\ref{subsec:soft-vs-hard}, we choose soft-sharing over hard-sharing because of the more expressive parameter sharing it provides to the model. Empirical results in Table.~\ref{table:ablation-studies} prove that soft-sharing method is statistically significantly better than hard-sharing with $p < 0.001$ in all metrics.\footnote{In the interest of space, most of the analyses are shown for CNN/DailyMail experiments, but we observed similar trends for the Gigaword experiments as well.}

\paragraph{Comparison of Different Layer-Sharing Methods}
We also conducted ablation studies among various layer-sharing approaches. Table~\ref{table:ablation-studies} shows results for soft-sharing models with decoder-only sharing (D1+D2; similar to~\newcite{Pasunuru2017TowardsIA}) as well as lower-layer sharing (encoder layer 1 + decoder layer 2, with and without attention shared). As shown, our final model (high-level semantic layer sharing E2+Attn+D1) outperforms these alternate sharing methods in all metrics with statistical significance ($p<0.05$).\footnote{Note that all our soft and layer sharing decisions were strictly made on the dev/validation set (see Sec.~\ref{subsec:training-details}).}

\paragraph{Quantitative Improvements in Entailment}
We employ a state-of-the-art entailment classifier~\cite{chen2017enhanced}, and calculate the average of the entailment probability of each of the output summary's sentences being entailed by the input source document. We do this for output summaries of our baseline and 2-way-EG multi-task model (with entailment generation). As can be seen in Table~\ref{table:entailment_analysis_results}, our multi-task model improves upon the baseline in the aspect of being entailed by the source document (with statistical significance $p<0.001$). Further, we use the Named Entity Recognition (NER) module from CoreNLP~\cite{manning2014corenlp} to compute the number of times the output summary contains extraneous facts (i.e., named entities as detected by the NER system) that are not present in the source documents, based on the intuition that a well-entailed summary should not contain unrelated information not followed from the input premise. We found that our 2-way MTL model with entailment generation reduces this extraneous count by $17.2\%$ w.r.t. the baseline. The qualitative examples below further discuss this issue of generating unrelated information.

\paragraph{Quantitative Improvements in Saliency Detection}
\label{para:saliency-detection}
For our saliency evaluation, we used the answer-span prediction classifier from~\citet{pasunuru2018multi} trained on SQuAD~\cite{rajpurkar2016squad} as the keyword detection classifier. We then annotate the ground-truth and model summaries with this keyword classifier and compute the \% match, i.e., how many salient words from the ground-truth summary were also generated in the model summary. The results are shown in Table~\ref{table:saliency_results_2}, where the 2-way-QG MTL model (with question generation) versus baseline improvement is stat. significant ($p<0.01$).
Moreover, we found $93$ more cases where our 2-way-QG MTL model detects 2 or more additional salient keywords than the pointer baseline model (as opposed to vice versa), showing that sentence-level question generation task is helping the document-level summarization task in finding more salient terms.

\begin{figure}[t]
\centering
\includegraphics[width=1.0\linewidth]{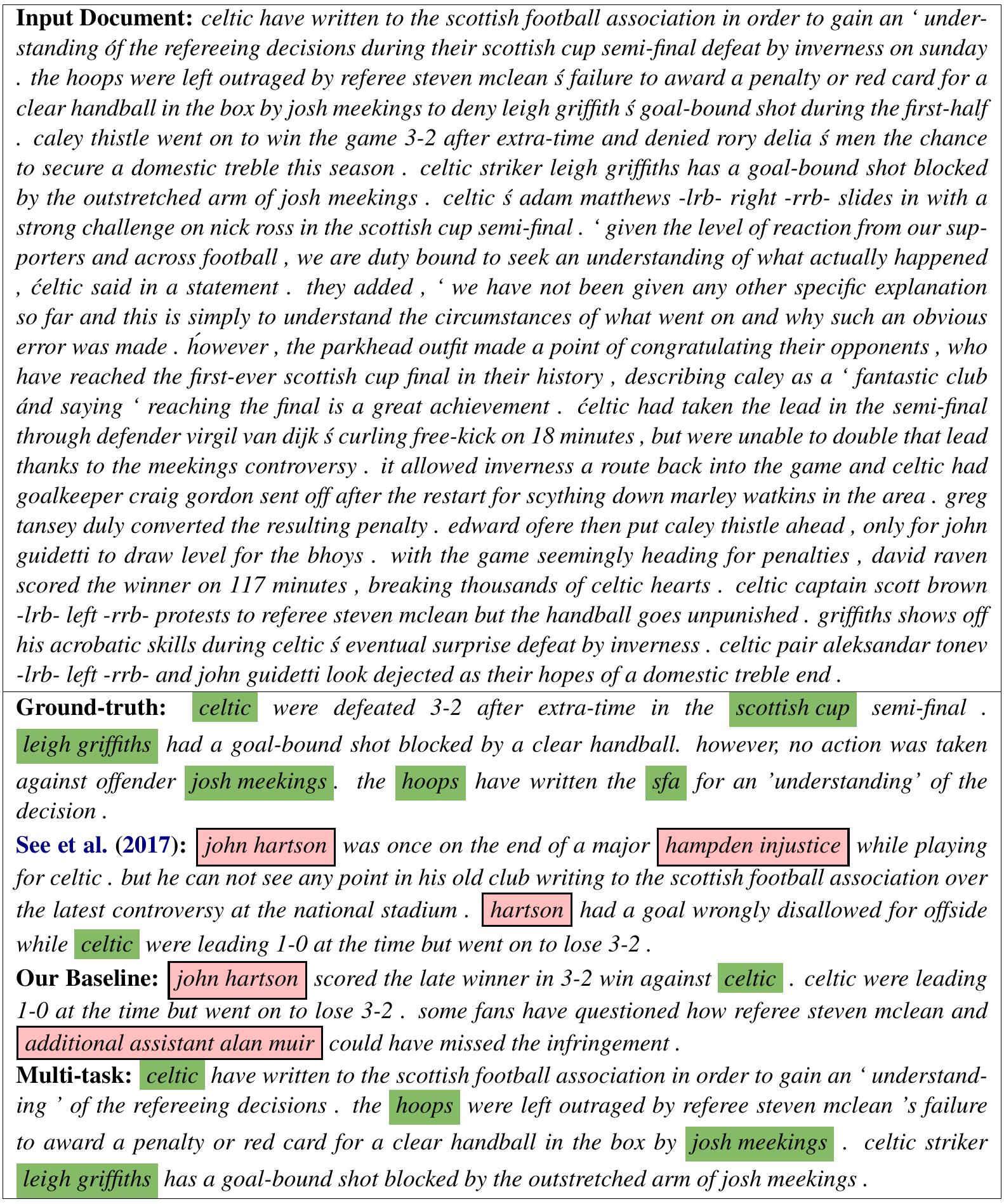}
\vspace{-26pt}
\caption{Example summary from our 3-way MTL model. The boxed-red highlights are extraneously-generated words not present/paraphrased in the input document. The unboxed-green highlights show salient phrases.
}
\label{fig:output-example}
\vspace{-6pt}
\end{figure}

\paragraph{Qualitative Examples on Entailment and Saliency Improvements}
\label{para:qualitative-analysis}
Fig.~\ref{fig:output-example} presents an example of output summaries generated by~\newcite{see2017get}, our baseline, and our 3-way multi-task model.~\newcite{see2017get} and our baseline models generate phrases like ``john hartson" and ``hampden injustice" that don't appear in the input document, hence they are not entailed by the input.\footnote{These extra, non-entailed unrelated/contradictory information are not present at all in any paraphrase form in the input document.} Moreover, both models missed salient information like ``josh meekings", ``leigh griffiths", and ``hoops", that our multi-task model recovers.\footnote{We consider the fill-in-the-blank highlights annotated by human on CNN/DailyMail dataset as salient information.} Hence, our 3-way multi-task model generates summaries that are both better at logical entailment and contain more salient information. We refer to supplementary Fig.~\ref{table:cnndm_entailment_saliency_examples} for more details and similar examples for separated 2-way multi-task models (supplementary Fig.~\ref{table:cnndm_entailment_examples}, Fig.~\ref{table:cnndm_saliency_examples}).

\begin{table}
\begin{small}
\begin{center}
\begin{tabular}{|l|c|c|c|}
\hline
Models & 2-gram & 3-gram & 4-gram \\
\hline
\newcite{see2017get} & 2.24 & 6.03 & 9.72 \\
MTL (3-way) & 2.84 & 6.83 & 10.66 \\
\hline
\end{tabular}
\end{center}
\vspace{-10pt}
\caption{Abstractiveness: novel n-gram percent.}
\label{table:abstractiveness_results}
\vspace{-8pt}
\end{small}
\end{table}

\paragraph{Abstractiveness Analysis}
As suggested in~\citet{see2017get}, we also compute the abstractiveness score as the number of novel $n$-grams between the model output summary and source document. As shown in Table~\ref{table:abstractiveness_results}, our multi-task model (EG + QG) is more abstractive than~\citet{see2017get}.

\section{Conclusion}
We presented a multi-task learning approach to improve abstractive summarization by incorporating the ability to detect salient information and to be logically entailed by the document, via question generation and entailment generation auxiliary tasks. We propose effective soft and high-level (semantic) layer-specific parameter sharing and achieve significant improvements over the state-of-the-art on two popular datasets, as well as a generalizability/transfer DUC-2002 setup. 

\section*{Acknowledgments}
We thank the reviewers for their helpful comments. This work was supported by DARPA (YFA17-D17AP00022), Google Faculty Research Award, Bloomberg Data Science Research Grant, and NVidia GPU awards. The views, opinions, and/or findings contained in this article are those of the authors and should not be interpreted as representing the official views or policies, either expressed or implied, of the funding agency.

\bibliography{acl2018}
\bibliographystyle{acl_natbib}

\appendix

\section{Supplementary Material}

\subsection{Dataset Details}
\label{sec:datasetsapp}
\paragraph{CNN/DailyMail Dataset}
CNN/DailyMail dataset~\cite{hermann2015teaching,nallapati2016abstractive} is a large collection of online news articles and their multi-sentence summaries. We use the original, non-anonymized version of the dataset provided by~\newcite{see2017get}. Overall, the dataset has $287,226$ training pairs, $13,368$ validation pairs and, $11,490$ test pairs. On an average, a source document has $781$ tokens and a target summary has $56$ tokens.

\paragraph{Gigaword Corpus}
Gigaword is based on a large collection of news articles, where the article's first sentence is considered as the input document and the headline of the article as output summary. We use the annotated corpus provided by~\newcite{rush2015neural}. It has around 3.8 million training samples. For validation, we use $2,000$ samples and for test evaluation we use the standard test set provided by~\newcite{rush2015neural}. Following previous work, we keep our vocabulary size to $50,000$ frequent words.

\paragraph{DUC Corpus}
We use the DUC-2002\footnote{\scriptsize{\url{https://www-nlpir.nist.gov/projects/duc/guidelines/2002.html}}} document summarization dataset for checking our model's generalizability capabilities. DUC-2002 corpus consists of $567$ documents with one or two human annotated reference summaries. We also tried beam retuning using DUC-2003\footnote{\scriptsize{\url{https://www-nlpir.nist.gov/projects/duc/guidelines/2003.html}}} as a validation set, which consists of $624$ documents with single human annotated reference summaries.

\paragraph{SNLI corpus}
We use the Stanford Natural Language Inference (SNLI) corpus~\cite{bowman2015large} for our entailment generation task. Following~\newcite{pasunuru2017multitask}, we use the same re-splits provided by them to ensure a zero train-test overlap and multi-reference setup. This dataset has a total of $145,822$ unique premise pairs out of $190,113$ pairs, which are used for training, and the rest of them are divided equally into validation and test sets.

\paragraph{SQuAD Dataset}
We use Stanford Question Answering Dataset (SQuAD) for our question generation task~\cite{rajpurkar2016squad}. In SQuAD dataset, given the comprehension and question, the task is to predict the answer span in the comprehension. However, in our question generation task, we extract the sentence from the comprehension containing the answer span and create a sentence-question pair similar to~\newcite{du2017learning}. The dataset has around $100$K sentence-question pairs from $536$ articles.

\begin{figure*}
\centering
\begin{tabularx}{\textwidth}{|X|l|}
\hline
\textbf{Input Document:} \emph{john hughes has revealed how he came within a heartbeat of stepping down from his job at inverness as the josh meekings controversy went into overdrive this week . the caley thistle boss says he felt so repulsed by the gut-wrenching predicament being endured by his young defender - before he was dramatically cleared - that he was ready to walk away from his post and the games he loves , just weeks before an historic scottish cup final date . keen cyclist hughes set off on a lonely bike ride after hearing meekings had been cited for the handball missed by officials in the semi-final against celtic , and admits his head was in a spin over an affair that has dominated the back-page headlines since last sunday . inverness defender josh meekings will be allowed to appear in scottish cup final after his ban was dismissed . only messages of support awaiting him on his return from footballing friends brought him back from the brink of quitting . hughes , who lives in the black isle just north of inverness , said : ` i came in here this morning after a day off . i turned my phone off and was away myself , away out on the bike with plenty of thinking time : a great freedom of mind . ` i was that sick of what has been going on in scottish football i was seriously contemplating my own future . i 'm serious when i say that . ` i had just had it up to here and was ready to just give it up . if it was n't for what happened when i turned my phone back on , with the phone calls and texts i received from people i really value in football , that my spirits picked up again . ` the calls and texts came in from all over the place , from some of the highest levels across the game . i 've had phone calls that have really got me back on my feet . ` i would n't like to name them all , but there were a lot of good people and a good few close friends in the football fraternity . meekings was not sent off and no penalty was given as inverness went on to beat celtic 3-2 after extra-time . ` they were saying : `` you need to lead from the front , you need to fight it . '' that restored and galvanised that focus and drive in me . and , if that was how i was feeling , how was the boy josh meekings feeling ? it should never have come to this . ' meekings was cleared to play in the final by the judicial panel yesterday , but hughes insists this ` unprecedented ' sfa wrangle must be the catalyst for change in scottish football 's governance . although those who sit on the panel are drawn from many walks of life , ranging from former players and coaches to ex-refs and members of the legal profession , hughes said he wants ` real football people ' drafted in instead of the ` suits ' he claims lack understanding of the nuances and spirit of the professional game . and he seemed to point a thinly-veiled finger of accusation at sfa chief executive stewart regan by alleging that compliance officer tony mcglennan was a mere ` patsy ' in the process . (...)}\\
\hline
\textbf{Ground-truth:} \emph{Inverness defender josh meekings has won appeal against one-match ban . the 22-year-old was offered one-game suspension following incident . however , an independent judicial panel tribunal overturned decision . inverness reached the scottish cup final with 3-2 win over celtic .} \\

\textbf{\newcite{see2017get}:} \emph{Josh meekings has been cleared to play in the scottish cup final .The englishman admitted he was fortunate not to have conceded a penalty and been sent off by \fboxrule=1pt\fcolorbox{black}{pink}{referee steven mclean} for stopping leigh griffiths’ net-bound effort on his goal-line . Meekings was not sent off and no penalty was given as inverness went on to beat celtic 3-2 .}\\

\textbf{Our Baseline:} \emph{Josh meekings cleared to play in the scottish cup final on \fboxrule=1pt\fcolorbox{black}{pink}{may 30} . Inverness defender josh meekings will be allowed to appear in scottish cup final . Meekings was not sent off and no penalty was given as inverness went on to beat celtic 3-2 .}\\

\textbf{Multi-task:} \emph{Josh meekings has been cleared to play in the scottish cup final . Inverness defender josh meekings will be allowed to appear in scottish cup final after his ban was dismissed . Inverness went on to beat celtic 3-2 after extra-time .}\\
\hline
\end{tabularx}

\caption{Example showing summaries generated by~\newcite{see2017get}, our baseline, and 2-way multi-task model with summarization and entailment generation. Boxed-red highlighted words/phrases are not present in the input source document in any paraphrasing form. As shown, both~\newcite{see2017get} and the baseline generate extraneous information that is not entailed by the source documents (``referee steven mclean'' and ``may 30''), but our multi-task model avoids such unrelated information to generate summaries that logically follow from the source document.}
\label{table:cnndm_entailment_examples}
\end{figure*}

\begin{figure*}
\centering
\begin{tabularx}{\textwidth}{|X|l|}
\hline
\textbf{Input Document:} \emph{bending and rising in spectacular fashion , these stunning pictures capture the paddy fields of south east asia and the arduous life of the farmers who cultivate them . in a photo album that spans over china , thailand , vietnam , laos and cambodia , extraordinary images portray the crop 's full cycle from the primitive sowing of seeds to the distribution of millions of tonnes for consumption . the pictures were taken by professional photographer scott gable , 39 , who spent four months travelling across the region documenting the labour and threadbare equipment used to harvest the carbohydrate-rich food . scroll down for video . majestic : a farmer wades through the mud with a stick as late morning rain falls on top of dragonsbone terraces in longsheng county , china . rice is a staple food for more than one-half the world 's population , but for many consumers , its origin remains somewhat of a mystery . the crop accounts for one fifth of all calories consumed by humans and 87 per cent of it is produced in asia . it is also the thirstiest crop there is - according to the un , farmers need at least 2,000 litres of water to make one kilogram of rice . mr gable said he was determined to capture every stage of production with his rice project - from the planting to the harvesting all the way down to the shipping of the food . after acquiring some contacts from experts at cornell university in new york and conducting his own research , he left for china last may and spent the next four months traveling . he said : ` the images were taken over a four month period from april to july last year across asia . i visited china , thailand , vietnam , laos and cambodia as part of my rice project . video courtesy of www.scottgable.com . breathtaking : a paddy field worker toils on the beautiful landscape of dragonsbone terraces in longsheng county , china . farmers ' procession : a rice planting festival parade takes place near the village of pingan in guangxi province , china . ` the project is one part of a larger three part project on global food staples - rice , corn and wheat . i am currently in the process of shooting the corn segment . ` the industrialisation of our food and mono-culture food staples have interested me for some time so that 's probably what inspired me to do this project . ' i shot the whole project using a canon slr and gopros . the actual shooting took four months and then post production took another four more months . ` the reaction to my work has been incredibly positive - i was able to secure a solo gallery show and create quite a bit of interest online which has been great . ' family crop : a hani woman in traditional clothing sits on top of her family 's rice store in yunnan province , china . arduous labour : employees of taiwan 's state-run rice experimental station are pictured beating rice husks by hand as the sun shines on them . mr gable spent months learning mandarin chinese in preparation for his trip , but the language barrier was still his greatest challenge . (...)
} \\
\hline
\textbf{Ground-truth:} \emph{the spectacular photos were taken at paddy fields in \colorbox{dollarbill}{china} , \colorbox{dollarbill}{thailand} , \colorbox{dollarbill}{vietnam} , \colorbox{dollarbill}{laos} and \colorbox{dollarbill}{cambodia} . photographer \colorbox{dollarbill}{scott gable} spent four months travelling region to document the process of harvesting the crop . \colorbox{dollarbill}{rice} accounts for one fifth of all calories consumed by humans but crop is often still cultivated in primitive way .} \\

\textbf{\newcite{see2017get}:} \emph{the pictures were taken by professional photographer \colorbox{dollarbill}{scott gable} , 39 , who spent four months travelling across the region documenting the labour and the arduous life of the farmers who cultivate them . the images were taken over a four month period from april to july last year across asia . mr gable said he was determined to capture every stage of production with his \colorbox{dollarbill}{rice} project .}\\

\textbf{Our Baseline:} \emph{\colorbox{dollarbill}{rice} is a staple food for more than one-half the world 's population . crop accounts for one fifth of all calories consumed by humans and 87 per cent of it is produced in asia .}\\

\textbf{Multi-task:} \emph{in a photo album that spans over \colorbox{dollarbill}{china} , \colorbox{dollarbill}{thailand} , \colorbox{dollarbill}{vietnam} , \colorbox{dollarbill}{laos} and \colorbox{dollarbill}{cambodia} , extraordinary images portray the crop 's full cycle from the primitive sowing of seeds to the distribution of millions of tonnes for consumption . the crop accounts for one fifth of all calories consumed by humans and 87 per cent of it is produced in asia .}\\
\hline
\end{tabularx}
\caption{Example showing summaries generated by~\newcite{see2017get}, our baseline, and 2-way multi-task model with summarization and question generation. All the unboxed-green highlighted words/phrases correspond to the salient information (based on the cloze-blanks of the original CNN/DailyMail Q\&A task/dataset~\cite{hermann2015teaching}). As shown, our multi-task model is able to generate most of this saliency information, while the outputs from~\newcite{see2017get} and baseline missed most of them, especially the country names.}
\label{table:cnndm_saliency_examples}
\end{figure*}

\begin{figure*}
\centering
\begin{tabularx}{\textwidth}{|X|l|}
\hline
\textbf{Input Document:} \emph{celtic have written to the scottish football association in order to gain an ` understanding \' of the refereeing decisions during their scottish cup semi-final defeat by inverness on sunday . the hoops were left outraged by referee steven mclean \'s failure to award a penalty or red card for a clear handball in the box by josh meekings to deny leigh griffith \'s goal-bound shot during the first-half . caley thistle went on to win the game 3-2 after extra-time and denied rory delia \'s men the chance to secure a domestic treble this season . celtic striker leigh griffiths has a goal-bound shot blocked by the outstretched arm of josh meekings . celtic \'s adam matthews -lrb- right -rrb- slides in with a strong challenge on nick ross in the scottish cup semi-final . ` given the level of reaction from our supporters and across football , we are duty bound to seek an understanding of what actually happened , \' celtic said in a statement . they added , ` we have not been given any other specific explanation so far and this is simply to understand the circumstances of what went on and why such an obvious error was made . \' however , the parkhead outfit made a point of congratulating their opponents , who have reached the first-ever scottish cup final in their history , describing caley as a ` fantastic club \' and saying ` reaching the final is a great achievement . \' celtic had taken the lead in the semi-final through defender virgil van dijk \'s curling free-kick on 18 minutes , but were unable to double that lead thanks to the meekings controversy . it allowed inverness a route back into the game and celtic had goalkeeper craig gordon sent off after the restart for scything down marley watkins in the area . greg tansey duly converted the resulting penalty . edward ofere then put caley thistle ahead , only for john guidetti to draw level for the bhoys . with the game seemingly heading for penalties , david raven scored the winner on 117 minutes , breaking thousands of celtic hearts . celtic captain scott brown -lrb- left -rrb- protests to referee steven mclean but the handball goes unpunished . griffiths shows off his acrobatic skills during celtic \'s eventual surprise defeat by inverness . celtic pair aleksandar tonev -lrb- left -rrb- and john guidetti look dejected as their hopes of a domestic treble end .} \\ 
\hline
\textbf{Ground-truth:} \emph{\colorbox{dollarbill}{celtic} were defeated 3-2 after extra-time in the \colorbox{dollarbill}{scottish cup} semi-final . \colorbox{dollarbill}{leigh griffiths} had a goal-bound shot blocked by a clear handball. however, no action was taken against offender \colorbox{dollarbill}{josh meekings}. the \colorbox{dollarbill}{hoops} have written the \colorbox{dollarbill}{sfa} for an 'understanding' of the decision .} \\

\textbf{\newcite{see2017get}:} \emph{\fboxrule=1pt\fcolorbox{black}{pink}{john hartson} was once on the end of a major \fboxrule=1pt\fcolorbox{black}{pink}{hampden injustice} while playing for celtic . but he can not see any point in his old club writing to the scottish football association over the latest controversy at the national stadium . \fboxrule=1pt\fcolorbox{black}{pink}{hartson} had a goal wrongly disallowed for offside while \colorbox{dollarbill}{celtic} were leading 1-0 at the time but went on to lose 3-2 .}\\

\textbf{Our Baseline:} \emph{\fboxrule=1pt\fcolorbox{black}{pink}{john hartson} scored the late winner in 3-2 win against \colorbox{dollarbill}{celtic} . celtic were leading 1-0 at the time but went on to lose 3-2 . some fans have questioned how referee steven mclean and \fboxrule=1pt\fcolorbox{black}{pink}{additional assistant alan muir} could have missed the infringement . }\\

\textbf{Multi-task:} \emph{\colorbox{dollarbill}{celtic} have written to the scottish football association in order to gain an ` understanding ' of the refereeing decisions . the \colorbox{dollarbill}{hoops} were left outraged by referee steven mclean 's failure to award a penalty or red card for a clear handball in the box by \colorbox{dollarbill}{josh meekings} . celtic striker \colorbox{dollarbill}{leigh griffiths} has a goal-bound shot blocked by the outstretched arm of josh meekings .}\\
\hline
\end{tabularx}
\caption{Example of summaries generated by~\newcite{see2017get}, our baseline, and 3-way multi-task model with summarization and both entailment generation and question generation. The boxed-red highlighted words/phrases are not present in the input source document in any paraphrasing form. All the unboxed-green highlighted words/phrases correspond to the salient information. See detailed discussion in Fig.~\ref{table:cnndm_entailment_examples} and Fig.~\ref{table:cnndm_saliency_examples} above. As shown, the outputs from~\newcite{see2017get} and the baseline both include non-entailed words/phrases (e.g. ``john hartson''), as well as they missed salient information (``hoops'', ``josh meekings'', ``leigh griffiths'') in their output summaries. Our multi-task model, however, manages to accomplish both, i.e., cover more salient information and also avoid unrelated information.
}
\label{table:cnndm_entailment_saliency_examples}
\end{figure*}

\subsection{Training Details}
The following training details are common across all models and datasets. 
We use LSTM-RNN in our sequence models with hidden state size of $256$ dimension. We use $128$ dimension word embedding representations. We do not use dropout or any other regularization techniques, but we clip the gradient to allow a maximum gradient norm value of $2.0$.
We use Adam optimizer~\cite{kingma2014adam} with a learning rate of $0.001$.
Also, we share the word embeddings representation of both encoder and decoder in our models. All our tuning decisions (including soft/hard and layer-specific sharing decisions) were made on the appropriate validation/development set.

\noindent\textbf{CNN/DailyMail}: For all the models involving CNN/DailyMail dataset, we use a maximum encoder RNN step size of $400$ and a maximum decoder RNN step size of $100$. We use a mini-batch size of $16$. We initialize the LSTM-RNNs with uniform random initialization in the range $[-0.02,0.02]$. We set $\lambda$ to $1.0$ in the joint cross-entropy and coverage loss. Also, we only add coverage to the converged model with attention and pointer mechanism, and make the learning rate from $0.001$ to $0.0001$. During multi-task learning, we use coverage mechanism for primary (CNN/DailyMail summarization) task but not for auxiliary tasks (because they do not have traditional redundancy issues). The penalty coefficient $\gamma$ for soft-sharing is set to $5\times 10^{-5}$ and $1\times 10^{-5}$ for 2-way and 3-way multi-task models respectively (the range of the penalty value is intuitively chosen such that we balance the cross-entropy and regularization losses). 
In inference time, we use a beam search size of $4$, following previous work~\cite{see2017get}. 

\noindent\textbf{Gigaword}: For all the models involving Gigaword dataset, we use a maximum encoder RNN step size of $50$ and a maximum decoder RNN step size of $20$. We use a mini-batch size of $256$. We initialize the LSTM-RNNs with uniform random initialization in the range $[-0.01,0.01]$. We do not use coverage mechanism to our Gigaword models. Also, we set our beam search size to $5$, following previous work~\cite{nallapati2016abstractive}.

\noindent\textbf{DUC}: For the CNN/DM to DUC domain-transfer experiments where we allow the beam sizes of all models to be individually re-tuned on DUC-2003, the chosen tuned beam values are $10, 4, 3$ for the multi-task model, baseline, and \newcite{see2017get}, respectively.

\begin{figure}
\begin{center}
\begin{small}
\begin{tabularx}{\linewidth}{|X|l|}
\hline
\textbf{Premise:} \emph{People walk down a paved street that has red lanterns hung from the buildings.} \\ 
\textbf{Entailment:} \emph{People walk down the street.} \\ 
\hline
\textbf{Premise:} \emph{A young woman on a boat in a light colored bikini kicks a man wearing a straw cowboy hat.}\\
\textbf{Entailment:} \emph{A young woman strikes a man with her feet.} \\ 
\hline
\end{tabularx}
\end{small}
\end{center}
\vspace{-9pt}
\caption{Output examples from our entailment generation model.}
\label{fig:entailment_examples}
\end{figure}

\begin{figure}[t]
\begin{center}
\begin{small}
\begin{tabularx}{\linewidth}{|X|l|}
\hline
\textbf{Input:} \emph{The college of science was established at the university in 1865 by president father patrick dillon.}\\ 
\textbf{Question:} \emph{In what year was the college of science established ?}\\
\hline
\textbf{Input:} \emph{Notable athletes include swimmer sharron davies , diver tom daley , dancer wayne sleep , and footballer}\\
\emph{trevor francis .} \\ 
\textbf{Question:} \emph{What is the occupation of trevor francis ? } \\ 
\hline
\end{tabularx}
\end{small}
\end{center}
\vspace{-9pt}
\caption{Output examples from our question generation model.}
\label{fig:question_examples}
\end{figure}
\subsubsection{Multi-Task Learning Details}

\paragraph{Multi-Task Learning with Question Generation}
Two important hyperparameters tuned are the mixing ratio between summarization and entailment generation, as well as the soft-sharing coefficient.
Here, we choose the mixing ratios $3{:}2$ between CNN/DailyMail and SQuAD, $100{:}1$ between Gigaword and SQuAD. Intuitively, these mixing ratios are close to the ratio of their dataset sizes. 
We set the soft-sharing coefficient $\gamma$ to $5\times 10^{-5}$ and $1\times 10^{-5}$ for CNN/DailyMail and Gigaword, resp.
 
\paragraph{Multi-Task Learning with Entailment Generation}
Here, we choose the mixing ratios $3{:}2$ between CNN/DailyMail and SNLI, $20{:}1$ between Gigaword and SNLI. We again set the soft-sharing coefficient $\gamma$ to $5\times 10^{-5}$ and $1\times 10^{-5}$ for CNN/DailyMail and Gigaword, resp.

\paragraph{Multi-Task Learning with Question and  Entailment Generation}
 Here, we choose the mixing ratios and soft-sharing coefficients to be $4{:}3{:}3$ and $5\times 10^{-5}$ for CNN/DailyMail, and $100{:}1{:}5$ and $1.5\times 10^{-6}$ for Gigaword respectively. 

\subsection{Auxiliary Output Analysis}
\subsubsection{Entailment Generation Examples}
See~\figref{fig:entailment_examples} for interesting output examples by our entailment generation model.

\subsubsection{Question Generation Examples}
See~\figref{fig:question_examples} for interesting output examples by our question generation model.

\end{document}